%% file: tbn.tex
\def\year{2019}\relax
\documentclass[letterpaper]{article} %DO NOT CHANGE THIS
\usepackage{aaai19}  %Required
\usepackage{times}  %Required
\usepackage{helvet}  %Required
\usepackage{courier}  %Required
\usepackage{url}  %Required
\usepackage{graphicx}  %Required
\usepackage{microtype}
\usepackage{amsmath}
\usepackage{amssymb}
\usepackage{subcaption}
\usepackage{multirow}

\begin{document}
\title{Temporal Bilinear Networks for Video Action Recognition}
\author{Yanghao Li~~~~~Sijie Song~~~~~ Yuqi Li~~~~~ Jiaying Liu\thanks{Corresponding author. This work was supported by National Natural Science Foundation of China under contract No. 61772043 and Peking University – Tencent Rhino Bird Innovation Fund.}\\
Peking University~~~~~\\
{lyttonhao@gmail.com}~~{ssj940920@pku.edu.cn}~~
{liyuqi.ne@gmail.com}~~
{liujiaying@pku.edu.cn}
% {\tt\small \{lyttonhao, liujiaying\}@pku.edu.cn}~~~{\tt\small \{winsty, shijianping5000, xiaodi.hou\}@gmail.com}
}
\maketitle
\graphicspath{{figures/}}
%\rowcolors{2}{gray!25}{white}

\input{src/abstract.tex}

\input{src/introduction.tex}

\input{src/related.tex}

\input{src/proposed.tex}

\input{src/experiment.tex}

\input{src/conclusion.tex}

{
\bibliographystyle{aaai}
\bibliography{egbib}
}

\end{document}

%% file: src/abstract.tex
\begin{abstract}
Temporal modeling in videos is a fundamental yet challenging problem in computer vision. In this paper, we propose a novel Temporal Bilinear (TB) model to  capture the temporal pairwise feature interactions between adjacent frames. Compared with some existing temporal methods which are limited in linear transformations, our TB model considers explicit quadratic bilinear transformations in the temporal domain for motion evolution and sequential relation modeling. We further leverage the factorized bilinear model in linear complexity and a bottleneck network design to build our TB blocks, which also constrains the parameters and computation cost.  We consider two schemes in terms of the incorporation of TB blocks and the original 2D spatial convolutions, namely wide and deep Temporal Bilinear Networks (TBN). Finally, we perform experiments on several widely adopted datasets including Kinetics, UCF101 and HMDB51. The effectiveness of our TBNs is validated by comprehensive ablation analyses and comparisons with various state-of-the-art methods. 
\end{abstract}

%% file: src/introduction.tex
\begin{section}{Introduction}

Deep convolutional neural networks (CNNs)~\cite{alexnet,resnet} have witnessed the tremendous progress in computer vision over the past few years. CNNs have demonstrated their power in many visual tasks, from image classification~\cite{imagenet,resnet}, object detection~\cite{rcnn,faster-rcnn}, semantic segmentation~\cite{fcn} and video action recognition~\cite{simonyan2014two,c3d}. However, the progress of video action recognition is relatively much slower. One of the main challenges in this area is the modeling of both spatial appearance and temporal motion across different frames. Therefore, researchers in video action recognition recently devote most efforts to the effective modeling of temporal dynamics in the deep architectures.

There are three typical schemes of temporal motion modeling for deep learning methods in action recognition. (1) Some methods capture temporal dependencies by utilizing temporal pooling~\cite{karpathy2014large} or recurrent layers~\cite{ng2015beyond,karpathy2014large} on the top of the 2D CNNs. However, the visual features are extracted independently by CNNs in a frame-wise manner, and the recurrent layers are fed in merely one-dimensional high-level semantic features without spatial information. It makes the temporal dynamics ignored in the preceding CNNs, especially for subtle motion dynamics. (2) %Another way is to feed the network with optical flow as input at the beginning of 2D CNNs. These methods are based on two-stream networks~\cite{simonyan2014two,tsn} which capture appearance and motion information separately by different stream networks. 
Following the two-stream architecture~\cite{simonyan2014two,tsn}, many methods capture appearance and motion information separately by different stream networks with RGB and optical flow as input. Despite of the good results, it also potentially prevents the model from fully utilizing the appearance and motion information in a single network. Furthermore, the estimation of optical flow  is very time- and resource-consuming. Thus, they are unsuitable for existing large-scale datasets~\cite{kay2017kinetics}. (3) Different from the above methods which use 2D CNNs, methods like C3D~\cite{c3d} adopt 3D convolution operators to learn spatio-temporal structures directly from RGB frames. Recently, 3D CNNs start to present their effectiveness on some large-scale action recognition datasets~\cite{carreira2017quo}, but the temporal modeling in 3D convolution operators is still limited in linear transformation. It is unclear whether 3D convolution operators are effective enough to capture complex temporal relations across frames. In addition, 3D convolution operators also introduce much more parameters.

In this paper, we propose a new Temporal Bilinear (TB) model to enhance the capacities of CNNs to model spatio-temporal dependencies. Specifically, TB model employs bilinear transformation to capture pairwise interactions among CNN features of adjacent frames for video action recognition. %The TB model performs similar to convolutional layers but it employs the bilinear transformation instead of the linear transformation. 
We believe this explicit temporal quadratic transformations on adjacent frames are more powerful to model complex motion relations in the temporal domain. We insert the TB model into the original 2D CNNs to model appearance and motion information simultaneously. At the same time, through multiple TB blocks embedded in different layers of TBNs, multiple levels of temporal dynamics could be captured with different temporal receptive fields.

To avoid the explosion of quadratic parameters in bilinear models, we explore the factorized bilinear model~\cite{li2017factorized} in the temporal domain for video data. To further reduce the complexity of our TB model, we build the temporal bilinear block with a bottleneck structure, which stacks two temporal convolutional layers and one TB layer in between. %This makes the complexity of parameter and computation in TB block much lower than 3D convolution operators. 
The TB model could be built by existing common neural network layers, thus it could be trained effectively and seamlessly with the whole network.

To summarize, our contributions of this paper are three-fold:
\begin{itemize}
\item We present a novel Temporal Bilinear (TB) model to consider the temporal pairwise feature interactions across adjacent frames. By incorporating the TB model into 2D CNNs, original 2D convolution operators are complemented to capture both appearance context and motion dynamics. We also construct two different TBNs: Wide and Deep Temporal Bilinear Networks (WTBN and DTBN) to explore the effective combination of 2D convolution layers and the TB model.
\item The TB model is implemented based on factorized bilinear model with linear complexity. To further reduce the computation cost, we leverage the bottleneck design to build the TB block. Thus, the complexity of our model is much lower than 3D convolution operators in terms of both parameters and computation cost.
\item The effectiveness of our approach is validated on several standard benchmarks, including Kinetics~\cite{kay2017kinetics}, UCF101~\cite{soomro2012ucf101} and HMDB51~\cite{kuehne2011hmdb}. Our proposed method achieves superior or comparable results to state-of-the-art methods.
\end{itemize}

\end{section}

%% file: src/related.tex
\begin{section}{Related Work}
\begin{paragraph}{Deep learning for action recognition.} 
After the breakthrough of deep learning in image recognition~\cite{alexnet}, many research works start to apply deep neural networks in video action recognition~\cite{karpathy2014large,simonyan2014two,ji20133d,c3d}. In~\cite{karpathy2014large}, several temporal fusion strategies were explored when applying 2D CNNs on video data, but the performance is not satisfying compared to traditional hand-craft features~\cite{wang2013action}. Later, Long Short-Term Memory (LSTM) networks appended after CNNs were investigated for better sequence modeling in action recognition~\cite{ng2015beyond,donahue2015long}. 
%Although LSTM and CNN are end-to-end jointly trainable, the preceding CNNs still have no considerations about temporal information when modeling appearance information. 
High-level semantic features from CNNs are fed into the following recurrent layers. Thus, it is hard for the network to capture subtle motion dynamics across frames, even though LSTM and CNNs are jointly end-to-end trainable. 

In~\cite{simonyan2014two}, the two-stream architecture was proposed for action recognition, which contains one spatial stream fed with RGB data and one temporal stream taking optical flow as input. The final results of the two-stream networks are obtained by fusion of softmax scores. Based on the two-stream structure, Temporal Segment Network~\cite{tsn} further performed sparse sampling and temporal fusion to capture global structures in the videos, and achieved state-of-the-art results on UCF101~\cite{soomro2012ucf101} and HMDB51~\cite{kuehne2011hmdb}. To obtain powerful video-level representation with two-stream networks, ActionVLAD~\cite{girdhar2017actionvlad} incorporated learnable spatio-temporal aggregation methods in the networks. One dilemma of two-stream networks lies in the inefficient extraction of optical flow, especially for large-scale datasets~\cite{kay2017kinetics} and practical applications.

Another typical approach for CNN-based action recognition is 3D CNNs, which extend convolution operations into the temporal domain~\cite{ji20133d}. By incorporating 3D convolutional layers and 3D pooling layers, C3D~\cite{c3d} proposed a standard 3D CNN architecture for generic feature extraction. Based on a more powerful CNN architecture, I3D~\cite{carreira2017quo} inflated the 2D convolutional filters into 3D convolutions in the Inception~\cite{bn} architecture and achieved state-of-the-art results on the large-scale Kinetics dataset~\cite{kay2017kinetics}. The most obvious problem for 3D CNNs is that they inevitably bring in much more parameters. Therefore, methods like ${\rm F_{ST}CN}$~\cite{sun2015human} factorized 3D convolutional kernel with a 2D spatial kernel and a 1D temporal kernel to reduce parameters.

Instead of using optical flow as input in two-stream methods, our work directly learns spatio-temporal features from RGB frames similar to 3D CNNs. Without 3D convolutional filters, our proposed TB blocks directly capture temporal evolutions between adjacent frames. Compared with 3D convolution filters that use linear transformation across different frames, our TB model enhances the capacity of the network by modeling temporal pairwise interactions. Due to the factorization bilinear scheme and bottleneck structure design, there are much fewer parameters in our TBNs than 3D CNNs.
\end{paragraph}
\begin{paragraph}{Bilinear models.} 
%\noident\textbf{}
A method called Bilinear Pooling~\cite{lin2015bilinear} was introduced to first incorporate bilinear models with CNNs for fine-grained image recognition. Bilinear Pooling calculates a global bilinear descriptor by averaging pooling of outer product of the final convolutional layer. Since the dimension of bilinear descriptors could be very large, several methods were proposed to reduce this quadratic dimensionality. Compact Bilinear Pooling~\cite{gao2015compact} presented two approximation methods to obtain compact bilinear representations. Furthermore, Factorized Bilinear model~\cite{li2017factorized} was proposed as a generalized bilinear model which is extended to convolutional layers and meets linear complexity. In~\cite{wang2017sort}, a Second-Order Response Tranform approach was propoosed to append element-wise product to a two-branch network module. Different from the above bilinear models for image recognition, we apply bilinear models on the temporal domain for video data, aiming to improve pairwise motion relations and dependency learning between adjacent frames. 

Another related work is Spatiotemporal Pyramid Network~\cite{wang2017spatiotemporal} for the action recognition task. This model was based on the two-stream architecture, and utilized compact bilinear to fuse high-level spatial and temporal features extracted from CNNs independently. In our work, TB blocks are introduced to model temporal relations between adjacent frames, and be embedded at different levels of CNNs, which is more flexible. Meanwhile, our TB model are also combined with original 2D convolutional layers to jointly capture spatio-temporal structure for video action recognition.
\end{paragraph}
\end{section}

%% file: src/proposed.tex
\newcommand{\row}[2] {#1_{#2 \cdot}}
\newcommand{\col}[2] {#1_{\cdot #2}}
\begin{section}{Temporal Bilinear Networks}

In this section we describe our proposed method for video action recognition. First, we discuss the existing standard temporal modeling methods. Next, we elaborate the details of our proposed Temporal Bilinear model. Finally, we introduce our design of the Temporal Bilinear block and explain how we incorporate it into the current 2D CNNs.

\begin{subsection}{Temporal Modeling Methods}
Suppose we have the input features  (usually the filter responses of one layer in the network) of $T$ frames. For simplicity, here we assume the features from each time step is one-dimensional. And we denote them as $\{\mathbf{x}^1, ... , \mathbf{x}^i, ..., \mathbf{x}^T\}$ where $\mathbf{x}^i \in \mathbb{R}^C$ and $C$ is the feature dimension, we aim to aggregate these features in the temporal domain for temporal modeling. The output signals of the temporal modeling are defined as $\{\mathbf{y}^1, ... , \mathbf{y}^i, ... , \mathbf{y}^{T'}\}$ where $\mathbf{y}^i \in \mathbb{R}^C$ and $T'$ is the output temporal dimension. 
%Here we assume the features are one dimensional and omit the spatial dimensions since we only consider temporal domain.
In practice, each output feature corresponds to several consecutive input frames. In the following, we consider one output signal $\mathbf{y}$ and its corresponding input features which are centered at $\mathbf{x}^i$. Then the temporal modeling methods could be formulated as follows:
\begin{equation}\label{eq:temporal_model}
\begin{aligned}
\mathbf{y} &= g(\{\mathbf{x}^{i + j}, l(k) \leq j \leq r(k) \}), \\
l(k) & =  1 - \lfloor (k + 1) / 2 \rfloor, \quad r(k) =  \lfloor k / 2 \rfloor,
\end{aligned}
\end{equation}
where $k$ is the number of considered input frames (\emph{e.g.} the kernel size in the pooling and convolutional layers) and $g(\cdot)$ is the aggregation function. Next we discuss two standard temporal modeling methods for the aggregation function $g$.

\begin{paragraph}{Temporal Pooling.} 
A natural choice of aggregation function $g$ is temporal pooling~\cite{ng2015beyond,c3d}, which extends the traditional spatial pooling layers to temporal domain. The common pooling strategies could be max or average pooling:
\begin{equation}\label{eq:pooling}
\begin{aligned}
y_c = \max_{j = l(k)}^{r(k)}x_c^{i + j}, \quad y_c &= \frac{1}{k}\sum_{j = l(k)}^{r(k)} x_c^{i + j}, 
\end{aligned}
\end{equation}
where $y_c$ and $x_c^{i + j}$ are the $c$-th element of $\mathbf{y}$ and $\mathbf{x}^{i +j}$, respectively. Such pooling operations are easy to implement and fast to compute, but they ignore valuable implicit  relations between different frames.
\end{paragraph}

\begin{paragraph}{Temporal Convolution.}
Similar to temporal pooling, the temporal convolution~\cite{c3d,carreira2017quo} is extended from the spatial convolution operator. It performs learnable transformation on input frames as follows:
\begin{equation}\label{eq:convolution}
\begin{aligned}
y_c &= \sum_{j = l(k)}^{r(k)} \mathbf{W}_c^j\mathbf{x}^{i + j}, 
\end{aligned}
\end{equation}
where $\mathbf{W}_c^j$ is the weight matrix for the $c$-th output neuron. Temporal convolution learns the transformation within several adjacent frames. However, the expressiveness of such linear transformation is limited to model complex motion structures. % And the relations between frames, especially two adjacent frames, are not explicitly modeled. 
\end{paragraph}

These temporal modeling methods could also be combined with spatial domain operators and extended to 3D pooling or convolutional layers~\cite{c3d,carreira2017quo}.

\end{subsection}

\begin{subsection}{Temporal Bilinear Model}
The above existing temporal modeling methods are lack of the capacity to explicitly capture interactions  between adjacent frames. We are motivated to exploit bilinear transformations by a novel Temporal Bilinear model for video action recognition.

\begin{paragraph}{Formulation.} 
Following the bilinear models in image recognition~\cite{lin2015bilinear,li2017factorized}, we define a generic temporal bilinear operation in deep neural networks as:
\begin{equation}\label{eq:bilinear}
\begin{aligned}
y_c &= {\mathbf{x}^{i}}^T\mathbf{W}_c\mathbf{x}^{i+1},
\end{aligned}
\end{equation}
where $\mathbf{W}_c \in \mathbb{R}^{C \times C}$ is the interaction weight matrix between the two adjacent frames. Since each time we only consider one output neuron $c$, we omit this subscript for simplicity.

Although the above bilinear model is capable of capturing the temporal interactions, it introduces a quadratic number of parameters in the weight matrix $\mathbf{W}$. Following the factorization scheme in~\cite{li2017factorized}, we adopt a factorized bilinear weight to reduce the computation cost and parameter complexity as follows:
\begin{equation}\label{eq:factorized}
\begin{aligned}
y &= {\mathbf{x}^{i}}^T\mathbf{F}^T\mathbf{F}{\mathbf{x}^{i+1}},
\end{aligned}
\end{equation} 
where $\mathbf{F} \in \mathbb{R}^{p \times C}$ is the factorized interaction weight between the $i$-th and $i$+1-th input frames with $p \in \mathbb{N}_0^+$ factors. The factor number $p$ constrains the complexity of the TB model. To explain the TB model more clearly, Eq.~(\ref{eq:factorized}) can be expanded as:
\begin{equation}
\begin{aligned}
y = \sum_{j=1}^{C}\sum_{k=1}^{C}\langle \col{\mathbf{f}}{j}, \col{\mathbf{f}}{k} \rangle x^i_jx^{i+1}_k,
\end{aligned}
\end{equation}
where $x^i_j$ and $x^{i+1}_k$ correspond to the $j$-th and $k$-th variables of the input features $\mathbf{x}^i$ and $\mathbf{x}^{i+1}$, $\col{\mathbf{f}}{j}$ is the $j$-th column of $\mathbf{F}$ and $\langle \col{\mathbf{f}}{j}, \col{\mathbf{f}}{k} \rangle$ calculates the inner product of $\col{\mathbf{f}}{j}$ and  $\col{\mathbf{f}}{k}$. Therefore, each pair of the variables between two adjacent frames has their own explicit interaction weight. Meanwhile, the shared $p$ factors also reduce the risk of overfitting in the bilinear model.

\end{paragraph}

\begin{figure}[t]
\centering
\centering
\includegraphics[width=0.7\linewidth]{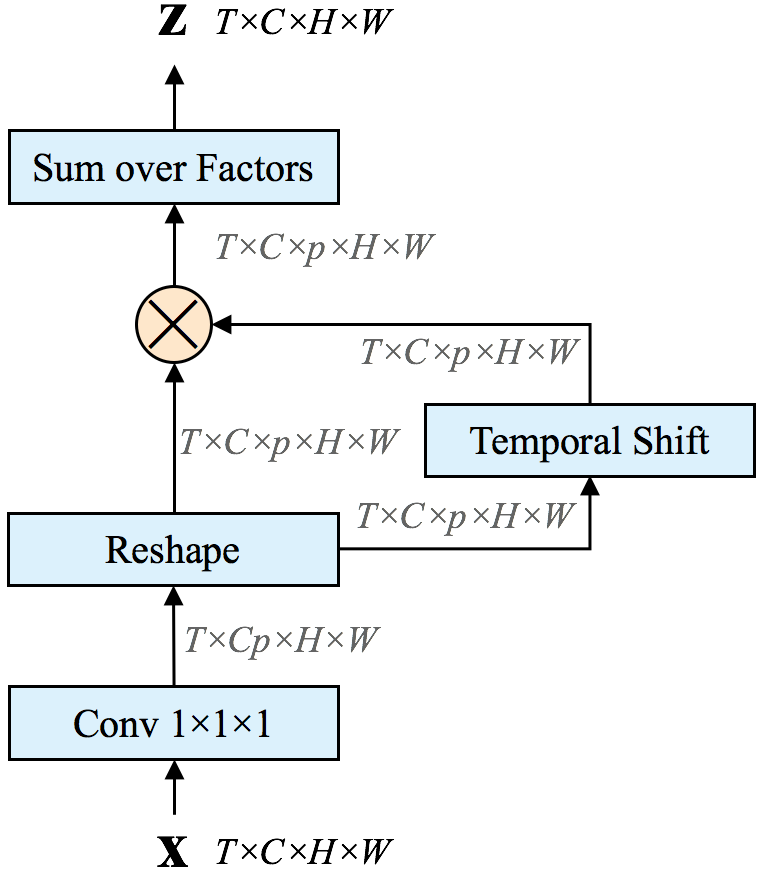}
\caption{The structure of Temporal Bilinear module. %Here the dimension of output $\mathbf{z}$ is set as the same with the input $x$. 
The feature maps are shown as the shape of their tensors. "$\otimes$" denotes element-wise multiplication.}\label{fig:TB_module}
\end{figure}

\begin{paragraph}{Instantiation.}

For video action recognition, the input feature $\mathbf{x}^i$ is usually three dimensional. To implement the above TB model in Eq.~(\ref{eq:factorized}) efficiently, we propose a general TB module for 3D feature maps using existing common neural network operators. Suppose the input feature map $\mathbf{x}$ is $T \times C \times H \times W$ where $C$ is the number of feature channel, $H$ and $W$ are the height and width of the feature map. Figure~\ref{fig:TB_module} shows an example of the TB module. Here  the output temporal dimension is the same as the input, \emph{i.e.}, $T' = T$.

The TB module is based on  Eq.~(\ref{eq:factorized}) and extended to combine with spatial domain. First, the convolution operator with $Cp$ filters calculates the transformation $\mathbf{F}\mathbf{x}^i$. Then we use a temporal shift operator, which could be implemented by some indexing operators, to create the tensor $\mathbf{F}\mathbf{x}^{i+1}$ where the time index starts from 2 (The last element is padded with the $T$-th frame). Finally, we utilize the element-wise multiplication and sum over the factor axis (the third axis) to obtain the final TB results ${\mathbf{x}^{i}}^T\mathbf{F}^T\mathbf{F}{\mathbf{x}^{i+1}}$ for each spatial and temporal element. Therefore, the TB model could be easily implemented and incorporated into the standard CNN architectures. Note that different from the factorization implementation in~\cite{li2017factorized}, our TB module is built on standard neural network operators in common deep learning platforms. Thus, it could be fully optimized by the standard optimization libraries such as cuDNN~\cite{cudnn}.
\end{paragraph}

\begin{table*}[htbp]
\small
	\begin{center}
	\begin{tabular}{l|c|c|c}
		\hline	
		Method							& Parameter		& Computation & Temporal RFS		\\
		\hline
		2D Conv ($3 \times 3$)				& $9C^2$		& O($9QC^2$)  &	1 \\
		3D Conv ($3 \times 3 \times 3$)		& $27C^2$ 		& O($27QC^2$) &	3\\
		TB Block						& $pC^2$ = $20C^2$		& O($pQC^2$)  = O($20QC^2$)	&	2 		\\
		Bottleneck TB Block				& $(6 + \frac{p}{16})C^2$ = $7.25C^2$	& O($(6 + \frac{p}{16})QC^2$)	= O($7.25QC^2$)	& 6		\\
		\hline
	\end{tabular}
	\end{center}
\caption{The comparisons with the proposed TB blocks and convolution operators. The value of $Q$ depends on the strides of spatial and temporal operators, \emph{e.g.}, $Q=HWT$ when stride is 1. The factor number $p$ is set as 20~\cite{li2017factorized}. Temporal RFS represents temporal Receptive Field Size.}\label{tb:comp_complexity}
\end{table*}

\begin{paragraph}{Temporal Bilinear Block.} 

\begin{figure}[t]
\begin{subfigure}{1.0\linewidth}
\centering
\includegraphics[width=0.54\linewidth]{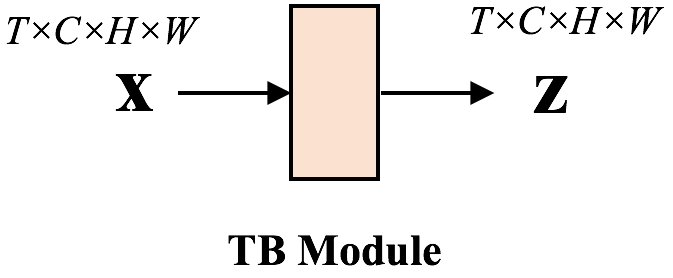}
\caption{TB Block}\label{fig:TB_block-a}
\end{subfigure}
\\
%\vspace{4mm}
\begin{subfigure}{1.0\linewidth}
\centering
\includegraphics[width=0.96\linewidth]{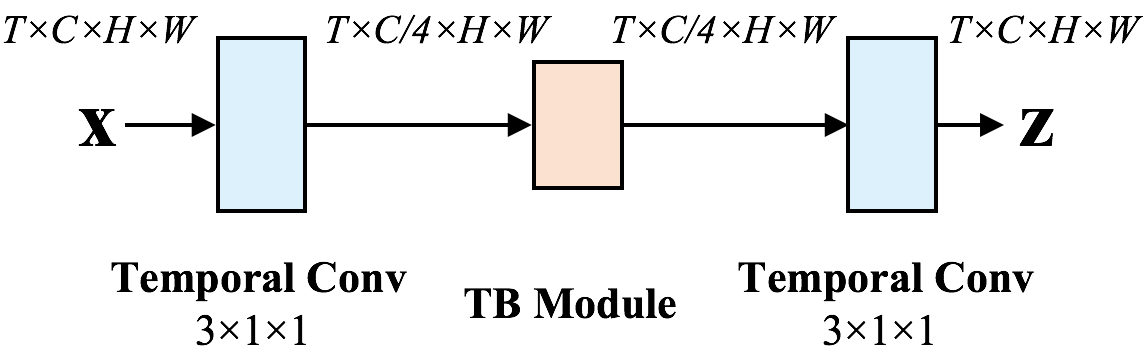}
\caption{Bottleneck TB Block}\label{fig:TB_block-b}
\end{subfigure}%
\caption{Two Temporal Bilinear Blocks. (a) Using one TB module. (b) Using Bottleneck structure.}\label{fig:TB_block}
\end{figure}

Although the complexity of TB model in Eq.~(\ref{eq:factorized}) is linear with the factor number $p$ and feature dimension $C$, it is still $p$ times larger than $1 \times 1$ convolution operators. Following the bottleneck design of~\cite{resnet}, we build a bottleneck Temporal Bilinear block to further reduce computation as shown in Figure~\ref{fig:TB_block}. Since our TB block focuses on temporal modeling, we append two temporal convolution operators before and after the TB module. The number of output channels of the first temporal convolution and the TB module are set as $C/4$. This reduces the computation of the TB module by 1/16. Table~\ref{tb:comp_complexity} compares the complexity and temporal Receptive Field Size (RFS) of the proposed TB blocks and 2D/3D convolution operators. As indicated, our bottleneck TB block reduces the complexity of parameters and computation by nearly 1/3, even lower than $3\times3$ 2D convolution operator. Further, the bottleneck TB block also achieves larger temporal RFS due to the stacked combination of temporal convolutions.

% \begin{figure}[t]
% \center{
% \subfigure[TB module]{\includegraphics[width=0.4\linewidth]{TB_block}\label{fig:TB_block-a} }~~
% \subfigure[Bottleneck TB Block] {\includegraphics[width=0.4\linewidth]{TB_bottleneck_block}\label{fig:TB_block-b}}~~
% }
% %\vspace{-1mm}
% \caption{Training on CIFAR-100 of Conv-FBN networks with $k = 20$ and different $p$. (a) Test error (\%) on CIFAR-100 of In5b-FBN and Conv-FBN networks. (b)(c) Training curves of In5b-FBN and Conv-FBN networks. Dashed lines denote training error, and bold lines denote testing error. Note that we do not show the results of smaller DropFactor rates, since the performance drops significantly when $p$ is too small. Best viewed in color.}\label{fig:result-drop}
% %\vspace{-2mm}
% \end{figure}
\end{paragraph}

\end{subsection}

\begin{subsection}{Temporal Bilinear Networks}
Our TB blocks are flexible to incorporate with standard CNNs. In this paper we adopt ResNet~\cite{resnet} owing to its good performance and simplicity. We first introduce the 2D ResNet baselines and then describe our proposed TBNs.
\begin{paragraph}{2D CNN Baseline.} 
\begin{table*}[hbtp]
%\scriptsize
%\small
%\renewcommand\arraystretch{1.15}
	\begin{center}

	\begin{tabular}{c|c|c|c|c}
		\hline	
		\multirow{2}{*}{layer name}							& \multicolumn{2}{c|}{C2D ResNet-18} & \multicolumn{2}{c}{C3D ResNet-18}			\\
		\cline{2-5}
		&	 layers & output & layers & output \\
		\hline
		conv1				& $7\times7$, 64, stride $2\times2$	& $8\times56\times56$ & $3\times7\times7$, 64, stride $1\times2\times2$	& $8\times56\times56$ \\
		\cline{1-3}\cline{4-5}
		res1 	& 
$\bigg[ \, \begin{aligned} 
& 3\times3, 64 \\
& 3\times3, 64  
\end{aligned} \, \bigg] \times 2$
 & $8\times56\times56$ & 
$\bigg[ \, \begin{aligned} 
& 3\times3\times3, 64 \\
& 3\times3\times3, 64  
\end{aligned} \, \bigg] \times 2$  & $8\times56\times56$ \\
		\cline{1-3}\cline{4-5}
		res2 	& 
$\bigg[ \, \begin{aligned} 
& 3\times3, 128 \\
& 3\times3, 128  
\end{aligned} \, \bigg] \times 2$
 & $8\times28\times28$ & 
$\bigg[ \, \begin{aligned} 
& 3\times3\times3, 128 \\
& 3\times3\times3, 128  
\end{aligned} \, \bigg] \times 2$  & $4\times28\times28$ \\
		\cline{1-3}\cline{4-5}
		res3 	& 
$\bigg[ \, \begin{aligned} 
& 3\times3, 256 \\
& 3\times3, 256  
\end{aligned} \, \bigg] \times 2$
 & $8\times14\times14$ & 
$\bigg[ \, \begin{aligned} 
& 3\times3\times3, 256 \\
& 3\times3\times3, 256  
\end{aligned} \, \bigg] \times 2$  & $2\times14\times14$ \\
		\cline{1-3}\cline{4-5}
		res4 	& 
$\bigg[ \, \begin{aligned} 
& 3\times3, 512 \\
& 3\times3, 512  
\end{aligned} \, \bigg] \times 2$
 & $8\times7\times7$ & 
$\bigg[ \, \begin{aligned} 
& 3\times3\times3, 512 \\
& 3\times3\times3, 512  
\end{aligned} \, \bigg] \times 2$  & $1\times7\times7$ \\
		\hline
&	global average pooling, fc & $1\times1\times1$ & global average pooling, fc & $1\times1\times1$ \\
		\hline
	\end{tabular}
	\end{center}
\caption{The architectures for C2D and C3D ResNet-18. The dimension of feature maps is $T\times H \times W$. The input size is $8\times112\times112$.  Residual blocks are shown in brackets and followed by the repeated numbers of blocks.}\label{tb:resnet_arch}
\end{table*}
In this paper, we adopt 2D ResNet and C3D ResNet as our baseline CNN structures. Table~\ref{tb:resnet_arch} shows the 2D ResNet-18 and C3D ResNet-18 structures. The input video clip consists of 8 frames with the resolution of $112 \times 112$. Note that the 2D kernels in ResNet-18 are equivalent to $1 \times k \times k$ kernels.

\end{paragraph}

\begin{paragraph}{Wide and Deep TBNs.} 
Our TB blocks are flexible to insert into both C2D and C3D ResNet. Since the TB blocks are responsible for temporal modeling, we mainly focus on incorporating them into C2D ResNet and compare the full model with C3D ResNet to validate its effectiveness in temporal domain. This also reduces the complexity of parameters and computation with C2D ResNet. To combine with spatial convolution operators in the residual block, we investigate parallel and serial integration schemes.

Figure~\ref{fig:TBN_blocks} shows the structures of the proposed Wide and Deep TB blocks in terms of the integration scheme. Note that we also add the identify path of the $3\times3$ Conv to keep the original appearance stream. Therefore, for the Wide TB block, the appearance and temporal information is learned in two parallel paths. While in the Deep TB block, temporal modeling is appended after the spatial convolutions. Finally, we replace the original block (Figure~\ref{fig:resnet_block}) in ResNet with our proposed TB blocks to construct Wide or Deep TBNs. 
\begin{figure}[h]
\center{
\begin{subfigure}{0.16\textwidth}
\centering
\includegraphics[height=3.4cm]{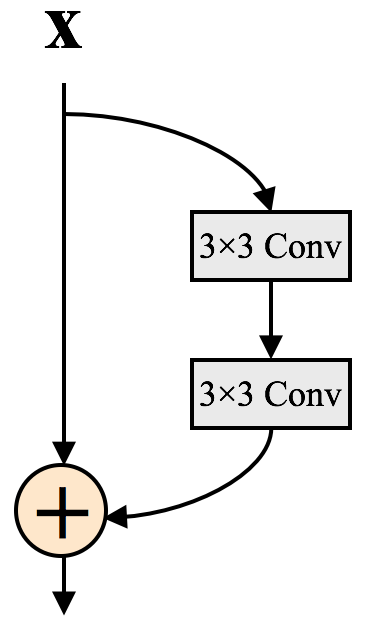}
\caption{ResNet block}\label{fig:resnet_block}
\end{subfigure}%
\begin{subfigure}{0.18\textwidth}
\centering
\includegraphics[height=3.4cm]{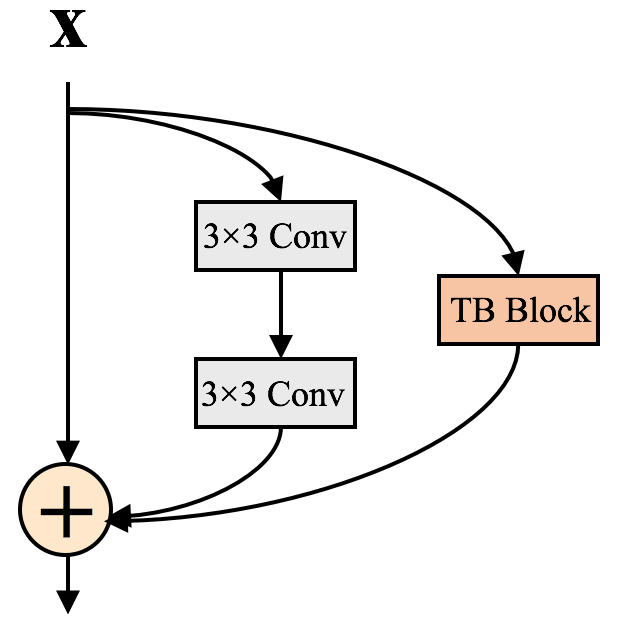}
\caption{Wide TB block}\label{fig:WTB_block}
\end{subfigure}%
\begin{subfigure}{0.16\textwidth}
\centering
\includegraphics[height=3.4cm]{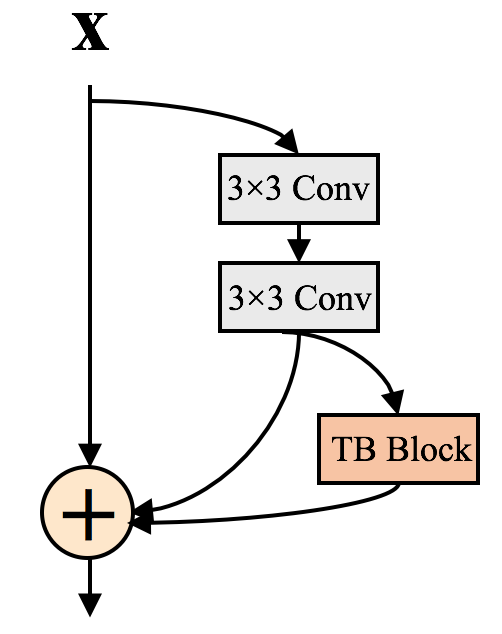}
\caption{Deep TB block}\label{fig:DTB_block}
\end{subfigure}%
% \subfigure[ResNet block]{\includegraphics[height= 4.4cm]{resnet_block}\label{fig:resnet_block} }~~
% \subfigure[Wide TB block] {\includegraphics[height= 4.4cm]{WTB_block}\label{fig:WTB_block}}~~
% \subfigure[Deep TB block] {\includegraphics[height= 4.4cm]{DTB_block}\label{fig:{DTB_block}}
% }
%\vspace{-1mm}
}
\caption{The structures of the original ResNet block, proposed Wide TB block and Deep TB block.}\label{fig:TBN_blocks}
%\vspace{-2mm}
\end{figure}
\end{paragraph}
\end{subsection}

\end{section}

%% file: src/experiment.tex
\begin{section}{Experiments}
In this section, we first conduct comprehensive ablation studies on Mini-Kinectis-200 dataset~\cite{xie2017rethinking}. Then we compare the results with state-of-the-art methods on Kinetics~\cite{kay2017kinetics}, UCF101~\cite{soomro2012ucf101} and HMDB51~\cite{kuehne2011hmdb} datasets. The proposed Wide and Deep TBNs are denoted by WTBN and DTBN, respectively.
\begin{paragraph}{Implementation Details.} 
Our models are trained on the training set of Kinetics dataset from scratch. All the network weights are initialized by the method in~\cite{he2015delving}. Following~\cite{tran2017convnet}, the networks take $8\times112\times112$ clips as input. The frame sampling stride is set as 4. The video frames are scaled to $128 \times 170$ and randomly cropped to $112\times112$. We train our models for 150 epochs with an initial learning rate of 0.1, which is decayed by a factor of 10 after 45, 90, 125 epochs. We use SGD as the optimizer with a weight decay of 0.0005 and batch size of 384. The standard augmentation methods like random cropping and random flipping are adopted during training for all the methods. For TBNs, we set the factor number $p$ as $20$ and also adopt the Dropfactor scheme~\cite{li2017factorized} to mitigate overfitting.

For testing, following the common evaluation scheme~\cite{tsn,tran2017convnet}, we uniformly sample 15 clips from input videos and then generate 10 crops for each clip. The final prediction results are obtained by averaging scores of  all the clips.
\end{paragraph}

\begin{subsection}{Ablation Study}
In this section, we investigate the design of TBNs with different ablation experiments. Since the full Kinetics dataset is quite large, we adopt the Mini-Kinetics-200~\cite{xie2017rethinking} for evaluation to speed up. It consists of 200 categories with most training examples from Kinetics. There are 80k and 5k videos in training and validation sets. For the baselines and TBNs, we utilize the ResNet-18 as our default backbone.

\begin{table*}[hbtp]
%\scriptsize
%\small
	\centering	
%\vspace{2mm}
	\begin{subtable}[t][][b]{0.32\linewidth}
		\centering
		\parbox[][4.0cm][t]{\linewidth}{% 手工指定高度、内容对齐
		\begin{tabular}{cc|cc}
		\hline
		Model & Stage & Top-1 & Top-5\\
		\hline
		C2D & - & 64.3 & 86.7 \\
		\hline
		\multirow{4}{*}{WTBN} 	& $\text{res}_1$ & 68.3 & 89.4 \\
							  	& $\text{res}_2$ & 69.0 & \textbf{89.8} \\
								& $\text{res}_3$ & \textbf{69.2}	& 89.4 \\
								& $\text{res}_4$ & 67.9 & 88.5 \\
		\hline
		\multirow{4}{*}{DTBN} 	& $\text{res}_1$ & \textbf{68.8} & 89.2 \\
							  	& $\text{res}_2$ & 68.6 & 89.3 \\
								& $\text{res}_3$ & \textbf{68.8} & \textbf{89.4} \\
								& $\text{res}_4$ & 66.9 & 88.5 \\
		\hline
		\end{tabular}}
		\caption{Stages}\label{table:ablation_stage}
	\end{subtable}
	\begin{subtable}[t][][b]{0.32\linewidth}
		\centering
		\parbox[][4.0cm][t]{\linewidth}{% 手工指定高度、内容对齐
		\begin{tabular}{cc|cc}
		\hline
		Model & Blocks & Top-1 & Top-5\\
		\hline
		C2D	& 0 & 64.3 & 86.7 \\
		\hline
		\multirow{3}{*}{WTBN} 	& 2 & 67.9 & 88.5  \\
							  	& 4 & 68.8 & 89.1 \\
								& 6 & \textbf{69.5} & \textbf{89.4} \\
		\hline
		\multirow{3}{*}{DTBN} 	& 2 & 66.9 & 88.5 \\
							  	& 4 & 68.3 & 89.4 \\
								& 6 & \textbf{69.0} & \textbf{89.6} \\
		\hline
		\end{tabular}}
		\caption{TB Blocks}\label{table:ablation_blocks}
	\end{subtable}
	\begin{subtable}[t][][b]{0.35\linewidth}
		\centering
		\parbox[][4.0cm][t]{\linewidth}{% 手工指定高度、内容对齐
		\begin{tabular}{c|c|cc}
		\hline
		Model & Params & Top-1 & Top-5\\
		\hline
		C2D	 & 11.3M & 64.3 & 86.7 \\
%		C2D-ResNet34 & 21.4M & 65.5 & 88.0 \\
		w/o bottleneck & 21.1M	&  66.6 & 87.8 \\
		w bottleneck & 12.9M	&  \textbf{68.8} & \textbf{89.1} \\

		\hline
		C3D	& 33.3M & 66.2 & 87.6 \\
		WTBN C2D & 11.4M & \textbf{69.0} & \textbf{89.8} \\
		WTBN C3D & 33.4M & 67.2 & 88.3 \\
		\hline
		\end{tabular}
		}
		\caption{Bottleneck Design $\&$ C3D TBN}\label{table:ablation_bottle_c3d}
	\end{subtable}
\caption{Ablations on Mini-Kinetics-200. The default backbone is ResNet-18. Evaluated by Top-1 and Top-5 accuracy (\%).}\label{table:ablation}
%\vspace{-2mm}
\end{table*}

\begin{table*}[hbpt]
%\small
% \vspace{-1mm}
	\begin{center}
	\begin{tabular}{lccc}
		\hline	
		Method		&	Backbone 	& Top-1 &  Top-5  \\
		\hline
		CNN+LSTM~\cite{kay2017kinetics} & ResNet-50 	& 57.0		& 79.0 \\
		RGB-Stream~\cite{kay2017kinetics} & ResNet-50 & 56.0		& 77.3 \\
		C3D~\cite{kay2017kinetics} & VGG-11 	& 56.1		& 79.5 \\
		I3D-RGB~\cite{carreira2017quo}	& Inception	& 68.4		&  88.0 \\
		\hline
		3D-Res~\cite{hara2017learning}  & ResNet-34 	& 58.0			& 81.3	\\
		TSN-RGB~\cite{tsn}	& Inception	& 69.1			&  88.7 \\
		\hline
		C2D/TBN	 & ResNet-18	& 61.1$\rightarrow$65.0		&  83.7$\rightarrow$86.4	 \\
		C2D/TBN  & ResNet-34	& 65.4$\rightarrow$69.5		& 86.4$\rightarrow$88.9		\\
		C2D/TBN  & ResNet-50	& 66.9$\rightarrow$\textbf{70.1}		& 87.2$\rightarrow$\textbf{89.3}	\\
		% TBN & ResNet-18	&   65.0	& 86.4 \\
		% TBN & ResNet-34	&	69.5 & 88.9 \\
		% TBN & ResNet-50 &	\textbf{70.1} & \textbf{89.3} \\
		% C2D/TBN	 & ResNet-18	& 61.1		&  83.7	 \\
		% C2D  & ResNet-34	& 65.4			& 86.4		\\
		% C2D  & ResNet-50	& 66.9		& 87.2	\\
		% TBN & ResNet-18	&   65.0	& 86.4 \\
		% TBN & ResNet-34	&	69.5 & 88.9 \\
		% TBN & ResNet-50 &	\textbf{70.1} & \textbf{89.3} \\
		% \hline
		% Backbone 	&	ResNet-50 	& ResNet-50 & VGG-11 	&	Inception	& ResNet-34 	& Inception	&	ResNet-18	&	ResNet-18	& ResNet-34	\\
		% Top-1 (\%)  & 	57.0		& 56.0		& 56.1		&	68.4		& 58.0			& 69.1			&	58.0		& 62.1	&	\textbf{69.5} \\
		% Top-5 (\%)  & 	79.0		& 77.3		& 79.5		&	88.0		& 81.3			& 88.7			& 	81.5		& 84.8	& 	\textbf{88.9} \\
		\hline
	\end{tabular}
	\end{center}
%\vspace{-1.5mm}
\caption{Comparisons on the validation set of Kinetics. Note that the first four methods are evaluated on testing set. Here we only compare the methods that use only RGB as input and are trained from scratch.} \label{tbl:kinetics}
%\vspace{-5mm}
\end{table*}

\begin{paragraph}{Stage to embed TB blocks.}
%\noindent\textbf{Stage to embed TB blocks.}
Table~\ref{table:ablation_stage} compares the results of  ResNet-18 with TB blocks embedded in different stages. We replace the two ResNet blocks in one stage with our Wide or Deep TB blocks. We can see that each TBN can lead to significant improvements (around 3\% to 5\%) on the C2D baseline for both WTBN and DTBN, which demonstrates the effectiveness of our TB model compared to the simple temporal pooling in C2D. The improvement of TB blocks on $\text{res}_4$ is relatively smaller. %The reason may be that the temporal motion information is insufficient for $\text{res}_4$ since it has small resolutions and is more related to high-level semantic features.
It is probably because $\text{res}_4$ is more related to high-level semantic features with insufficient temporal motion information. 
\end{paragraph}

\begin{paragraph}{Number of TB blocks.} We also investigate adding more TB blocks in TBN as shown in Table~\ref{table:ablation_blocks}. We add 2 ($\text{res}_4$), 4 ($\text{res}_4$, $\text{res}_3$) and 6 ($\text{res}_4$, $\text{res}_3$, $\text{res}_2$) TB blocks in WTBN and DTBN, respectively. Table~\ref{table:ablation_blocks} shows that more TB blocks lead to better results in general, which validates the capacity of temporal modeling of TBNs. It is also demonstrated that multiple TB blocks with larger temporal receptive fields perform better in modeling long-term temporal dependencies.
\end{paragraph}

\begin{paragraph}{Bottleneck design.} To validate the effectiveness of our bottleneck structure design in Figure~\ref{fig:TB_block}, we compare the results of TBNs with and without the bottleneck at the top of Table~\ref{table:ablation_bottle_c3d}. The two TBNs both use 4 TB blocks. The results demonstrate the bottleneck structure not only improves the performance (by 2.2\%) but also reduces the number of parameters a lot (by 39\%). Note that compared to C2D baselines, our bottleneck TBN achieves a significant improvement in performance with only 14\% additional parameters. 
\end{paragraph}

\begin{paragraph}{Combined with C3D ResNet.} In the above comparisons, the backbone network of TBNs is C2D ResNet. We further study the performance of adding TB blocks into C3D ResNet. The bottom of Table~\ref{table:ablation_bottle_c3d} shows the results of adding 2 TB blocks ($\text{res}_2$) into C2D and C3D, respectively. From the results, we can see that our WTBN C3D improves 1\% on C3D with almost equal number of parameters. It demonstrates that our TB blocks capture complementary information with 3D temporal convolutions. WTBN C3D does not achieve higher performance than WTBN C2D. %It may be caused by the temporal stride in WTBN C3D, which makes TB blocks not efficient enough.
It is mainly because of the gradual decrease of the temporal dimension in C3D, as shown in Table~\ref{tb:resnet_arch}, which weakens the capacity of temporal modeling in TB blocks.

\end{paragraph}

\begin{table*}[htbp]
%\small
%\vspace{-1mm}
	\begin{center}
	\begin{tabular}{lcccc}
		\hline	
		Method 		& Pretrain & backbone & UCF101	& HMDB51	\\
		\hline
		Two-Stream-RGB~\cite{simonyan2014two} 	& ImageNet	& VGG-M			& 73.0  & 40.5 \\
		TDD Spatial~\cite{wang2015action}		& ImageNet	& VGG-M			& 82.8	& 50.0 \\
		Res-RGB~\cite{feichtenhofer2016spatiotemporal} & ImageNet & ResNet-50 & 82.3 	& 43.4 \\
		TSN-RGB~\cite{tsn}						& ImageNet	& Inception		& \textbf{85.1}	& \textbf{51.0} \\
		I3D-RGB~\cite{kay2017kinetics}			& ImageNet	& Inception		& 84.5  & 49.8 \\
		\hline
		C2D										& ImageNet	& ResNet-18		& 76.9	& 41.2	\\
		TBN 									& ImageNet 	& ResNet-18		& 77.8	& 42.7	\\
		TBN 									& ImageNet	& ResNet-34		& \textbf{81.4}	& 	\textbf{46.4}	\\
		\hline	
		\hline
		C3D~\cite{c3d}					& Sports-1M				& VGG-11	& 82.3	& 51.6 \\
		TSN-RGB~\cite{tsn}				& ImageNet+Kinetics		& Inception	& 91.1	& - \\
		I3D-RGB~\cite{carreira2017quo} 	& ImageNet+Kinetics 	& Inception	& \textbf{95.6}  & \textbf{74.8} \\
		\hline
		C2D							    & Kinetics 				& ResNet-18	& 85.0	& 53.9 \\
		TBN 							& Kinetics          	& ResNet-18 & 89.6 	& 62.2 \\
		TBN 							& Kinetics              & ResNet-34 & \textbf{93.6}		& \textbf{69.4} \\
		\hline
	\end{tabular}

	\end{center}	
	\caption{Comparisons with state-of-the-art methods on UCF101 and HMDB51. The Top-1 accuracy (\%) is reported over 3 splits. Note we only consider methods that use only RGB as input.}\label{tbl:ucf_hmdb}
%\vspace{-5mm}
\end{table*}

\end{subsection}

\begin{subsection}{Evaluation on Multiple Datasets}
In this section, we compare our TBNs with other methods on multiple datasets, including Kinetics, UCF101 and HMDB51. Since the performance of WTBN and DTBN is slightly different, here we adopt WTBN with ResNet-18, ResNet-34 and ResNet-50 as our backbone networks. We add 6 TB blocks (in $\text{res}_2$, $\text{res}_3$ and $\text{res}_4$) for ResNet-18, 5 TB blocks (2 in $\text{res}_2$ and 3 in $\text{res}_3$) for ResNet-34, and 2 TB blocks (2 in $\text{res}_3$) For ResNet-50 in WTBN. 

\begin{subsubsection}{Results on Full Kinetics Dataset}
Kinetics~\cite{kay2017kinetics}  is a large-scale video action recognition dataset, which contains around 240k training videos and 20k validation videos with 400 action classes. %Due to its large scale, we compares the results of different methods with only RGB as input.

Table~\ref{tbl:kinetics} shows the results compared to some state-of-the-art methods. For a fair comparison, we consider the methods that only use RGB as input and are trained from scratch. Our proposed TBNs significantly improve the baseline methods. Meantime, our TBN with ResNet-18 achieves comparable results with C2D using ResNet-34 as backbone which has nearly twice the number of parameters. It demonstrates that the improvement of TBN is not just increasing depth and it is complementary to using deeper network. Compared to the recent state-of-the-art I3D~\cite{carreira2017quo} and TSN~\cite{tsn},% which use the superior Inception as their backbone networks, 
our method achieves the best Top-1 accuracy (70.1\%). Note that current published state-of-the-art methods could achieve higher performance, like~\cite{bian2017revisiting}, by utilizing more modalities, larger spatial resolutions and deeper structures.

% \begin{table*}[htbp]
% \scriptsize
% \vspace{-1mm}
% 	\begin{center}
% 	\setlength{\tabcolsep}{2pt}
% 	\begin{tabular}{l|cccc|cc|ccc}
% 		\hline	
% 		Method		&	CNN+LSTM~\cite{kay2017kinetics} & RGB-Stream~\cite{kay2017kinetics} & C3D~\cite{kay2017kinetics} & I3D-RGB~\cite{carreira2017quo}	& 3D-Res~\cite{hara2017learning}  & TSN-RGB~\cite{tsn}	& C2D	 & TBN & TBN 	\\
% 		\hline
% 		Backbone 	&	ResNet-50 	& ResNet-50 & VGG-11 	&	Inception	& ResNet-34 	& Inception	&	ResNet-18	&	ResNet-18	& ResNet-34	\\
% 		Top-1 (\%)  & 	57.0		& 56.0		& 56.1		&	68.4		& 58.0			& 69.1			&	58.0		& 62.1	&	\textbf{69.5} \\
% 		Top-5 (\%)  & 	79.0		& 77.3		& 79.5		&	88.0		& 81.3			& 88.7			& 	81.5		& 84.8	& 	\textbf{88.9} \\
% 		\hline
% 	\end{tabular}
% 	\end{center}
% %\vspace{-1.5mm}
% \caption{Comparisons on the validation set of Kinetics. Note that the first four methods are evaluated on testing set. Here we only compare the methods that use only RGB as input and are trained from scratch.} \label{tbl:kinetics}
% \vspace{-5mm}
% \end{table*}

\end{subsubsection}

\begin{subsubsection}{Results on UCF101 and HMDB51 Datasets}

We transfer the learned TBN models to two widely adopted action recognition datasets: UCF101~\cite{soomro2012ucf101} and HMDB51~\cite{kuehne2011hmdb}. UCF101 contains around 13,320 videos with 101 action classes, while HMDB51 has 6,766 videos from 51 action categories. We use the models trained on Kinetics or ImageNet as initialization and report the averaged accuracy over three splits. For finetuning, we use the same settings as Kinetics, but change the learning rate to 0.001 with total 100 training epochs.% and decay it at 50 and 75 epochs.

The results are summarized in Table~\ref{tbl:ucf_hmdb}. Our TBN consistently outperforms the baseline C2D method, regardless of the employed datasets in pretraining. It is observed that better results are obtained with ResNet-34. And the performance is further improved when pretrained with Kinetics (\emph{e.g.}, from 81.4\% to 93.6\%  on UCF101 for TBN with ResNet-34), owing to its large-scale and high-quality video data.
%First, no matter whether the models are pretrained on ImageNet or Kinetics, our TBN with ResNet-18 outperforms the baseline C2D method and using ResNet-34 as backbone further gets better results. Owing to the large scale and better quality of Kinetics, adopting Kinetics dataset for pretraining significantly improves performance (\emph{e.g.}, from 81.4\% to 93.6\%  on UCF101 for TBN with ResNet-34). 
Finally, our TBN obtains comparable performance with other state-of-the-art methods like TSN~\cite{tsn} and I3D~\cite{carreira2017quo} which adopt deeper network structures pretrained on ImageNet and Kinetics.

\end{subsubsection}

\end{subsection}
\end{section}

%% file: src/conclusion.tex
\begin{section}{Conclusions}
In this paper, we have presented the Temporal Bilinear (TB) model to incorporate temporal pairwise interactions in neural networks. %Thanks to the factorized bilinear model and bottleneck design, the TB model has low cost in both parameters and computation, and can be easily combined with existing 2D or 3D CNNs.
The factorized bilinear model and the bottleneck design bring fewer parameters and lower computational complexity. Besides, our TB block is very compact and flexible to combine with existing 2D or 3D CNNs.
The Temporal Bilinear Networks (TBN) achieve consistent improvements over baselines in several video action recognition benchmarks. We believe that TB model can be an essential component for temporal modeling and we will make efforts to apply TBNs to other video domain tasks in the future.
\end{section}